\documentclass[10pt]{article} 
\usepackage[preprint]{tmlr}

\usepackage{amsmath,amsfonts,bm}

\def\eqref#1{equation~\ref{#1}}

\def\1{\bm{1}}

\DeclareMathAlphabet{\mathsfit}{\encodingdefault}{\sfdefault}{m}{sl}
\SetMathAlphabet{\mathsfit}{bold}{\encodingdefault}{\sfdefault}{bx}{n}

\usepackage{microtype}
\usepackage{hyperref}
\usepackage{url}
\usepackage{booktabs}
\usepackage{graphicx}
\usepackage{float}
\usepackage{xcolor}
\usepackage{amsmath}
\usepackage{amssymb}
\usepackage{multirow}
\usepackage[font=small]{caption}
\usepackage[ruled,lined]{algorithm2e}

\definecolor{darkblue}{rgb}{0, 0, 0.5}
\hypersetup{colorlinks=true, citecolor=darkblue, linkcolor=darkblue, urlcolor=darkblue}

\title{Aspect-wise Pairwise Tournaments for Test-Time Selection}

\author{\name Hannah Guan \email hguan@college.harvard.edu \\
      \addr Harvard University
      \AND
      \name Yilun Du \email yilundu@seas.harvard.edu \\
      \addr Harvard University}

\def\month{MM}  
\def\year{YYYY} 
\def\openreview{\url{https://openreview.net/forum?id=XXXX}} 

\begin{document}

\maketitle

\begin{abstract}
Scaling test-time compute has proven highly effective at improving LLM reasoning. A popular method is best-of-$n$ sampling, where a model generates many candidate solutions and one is selected. This motivates the question: how can we select the best candidate? When humans make choices, we find it more natural to compare two options than judge each in isolation. Additionally, when the options become complex, we like to judge the options one aspect at a time, such as comparing two job candidates on experience, communication, etc. In the test-time setting, candidate solutions should be judged in the same way. Here we introduce \textbf{Aspect-wise Pairwise Tournaments (APT)}, which verifies by comparison and compares by aspect. Each matchup of a single elimination tournament is decided by a vote over multiple off-the-shelf aspect verifiers, each expressing a pairwise preference along its own aspect. Across five domains and three generator LLMs, APT improves over binary multi-aspect approval, self-consistency and reward model verification, with the largest gains on open-ended tasks. We find that both components matter; neither binary approval over the same aspects nor a generic pairwise judge at a matched budget matches APT. Because aspects are specified in natural language, APT also lets users steer selection toward properties they care about, better aligning with user preferences without needing to regenerate solutions. No component is trained, so APT transfers to new domains by simply rewriting prompts.
\end{abstract}

\section{Introduction}

Best-of-$n$ sampling, where, given a task, a model generates $n$ solutions and returns only the best one, substantially improves LLM performance without additional training \citep{cobbe2021training, lightman2023lets}. With current LLMs, generating many candidates is cheap, but it is often hard to identify the ``best" one. Many problems arise in the selection process, such as how to define ``best" or detect subtle differences in candidates. In this paper, we apply intuitions from human decision-making to develop a verification method. 

People rarely rank options by scoring them in isolation. An optometrist never directly asks a patient what their prescription should be. Instead, they show the patient a sequence of lenses side-by-side and ask, ``is one or two better?'' It's hard to have a concrete standard for what ``good" is, especially if the problem is not in your domain of expertise, as in the optometrist example. On the other hand, comparison draws attention to where two options differ, allowing an easier judgment to be made. The same can be said about LLM verifiers. Current models are not experts in every field, and two solutions might each sound plausible when read alone, but the differences that determine right from wrong come to light when they are read side by side.

Even comparison, though, gets harder as the options become more complex. A hiring committee choosing between two finalists does not ask which one is better overall. It compares them on technical depth, then on communication, then on experience, and only then weighs the results. This allows the committee to develop a definition of ``better", thus making the comparison more tractable. Different conditions also fail in different places, so an LLM verifier that compares along several axes is robust to a mistake on one of them.

\begin{figure}[t]
    \centering
    \includegraphics[width=\linewidth]{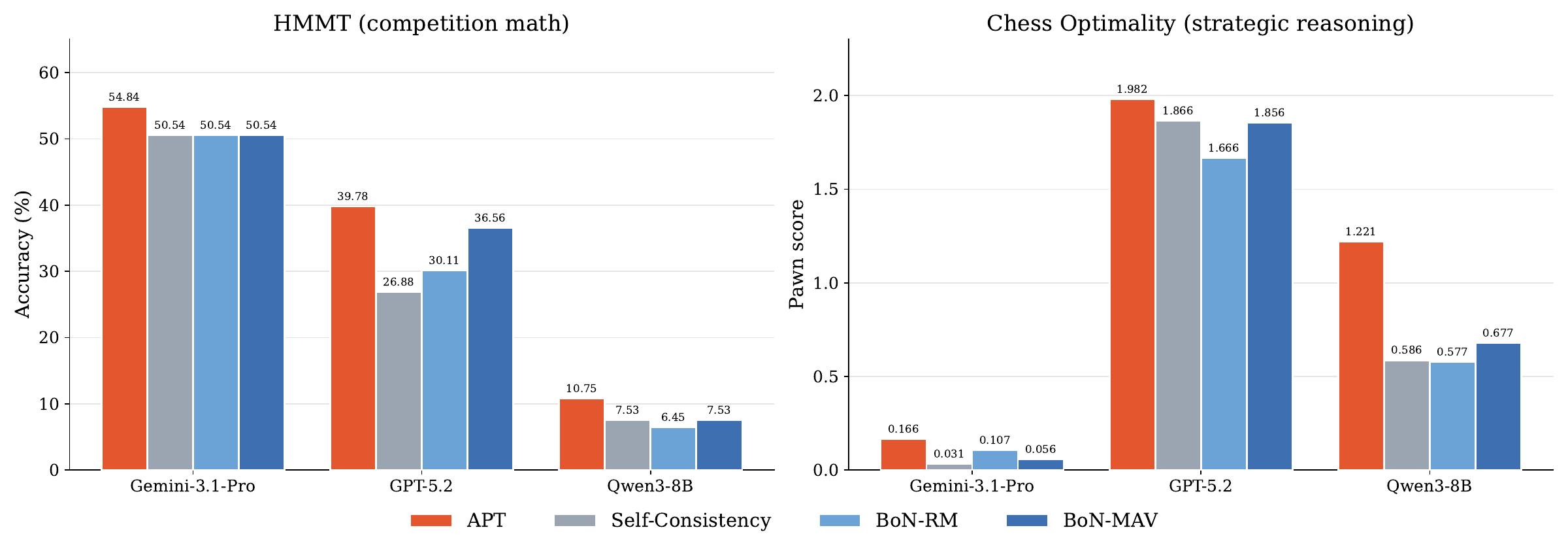}
    \vspace{-10pt}
    \looseness=-1
    \caption{APT improves test-time selection across models and domains. Comparison of Aspect-wise Pairwise Tournaments (APT), best-of-$n$ binary approval verification (BoN-MAV), best-of-$n$ reward model verification (BoN-RM), and self-consistency using $n = 8$ candidate solutions across three generator LLMs on two benchmarks. \textbf{Left:} Accuracy (\%) on HMMT competition mathematics. \textbf{Right:} pawn score on Chess Optimality dataset (higher is better). Full results are in Table~\ref{tab:main-results}.}
    \label{fig:teaser}
    \vspace{-12.5pt}
\end{figure}

Identifying a set of aspects also allows the committee to prioritize those they care most about. For instance, a committee hiring for a client-facing role will count communication more heavily than one hiring for a research role, even when looking at the same two candidates. What counts as ``better'' depends on each person's preferences and needs. With a single holistic judgment, there is only one notion of quality. If the axes are written down, a user can add one they care about and decide how much it should be weighted, thereby changing which candidate is selected without generating new ones.

Applying these observations on human decision-making to test-time verification, we introduce \textbf{Aspect-wise Pairwise Tournaments (APT)}. It verifies by comparison and compares by aspect. APT samples $n$ candidate solutions from a generator LLM and runs a single elimination tournament over them. Each matchup is decided by a vote across $m$ off-the-shelf aspect verifiers, each stating which of the two candidates it prefers along its own aspect. 

We evaluate APT on competition math, a hard multiple-choice benchmark, and two chess reasoning tasks, using three generator LLMs. APT achieves the best result in 14 of 15 generator-domain pairs against self-consistency, reward model verification, and binary multi-aspect approval (Section~\ref{sec:main-results}). Adding a weighted aspect steers selection toward a user's preference, selecting concise solutions on the math benchmarks and aggressive moves on Chess Optimality (Section~\ref{sec:steerability}). Both replacing the aspect vote with a generic comparative judge and binary approval over the same aspects fall short (Section~\ref{sec:ablations}).

Our contributions are as follows:
\begin{enumerate}
    \item \textbf{Aspect-wise Pairwise Tournaments.} A training-free test-time selection method that decides each matchup of a knockout tournament by a vote over aspect-specific pairwise verifiers.
    \item \textbf{Steering selection toward user preferences.} Because aspects are natural-language prompts, users can add and weight aspects at inference time to change what is selected, without regenerating candidates or training verifiers.
    \item \textbf{Isolating each component.} BoN-MAV~\citep{lifshitz2025multiagentverificationscalingtesttime}, which uses the same aspect verifiers with binary approval, isolates the contribution of pairwise comparison; a pairwise tournament with a generic judge at matched budget isolates the contribution of aspect decomposition. We find that both contributions improve accuracy.
\end{enumerate}

\section{Related Work}
\label{sec:related}

\textbf{Test-time compute scaling.}
Increasing inference-time computation can improve LLM performance without changing model parameters~\citep{snell2024scaling, wu2025inference, zhang2025survey}. As larger sample pools become more likely to contain a correct answer~\citep{brown2024large}, selecting among candidates becomes increasingly important. Existing approaches include best-of-$n$ selection with reward models~\citep{stiennon2020learning, cobbe2021training}, self-consistency for structured and free-form answers~\citep{wang2023selfconsistency, chen2023universal}, and search over intermediate reasoning steps~\citep{yao2023tree, lightman2023lets}. \citet{welleck2024decoding} survey these approaches as meta-generation. However, additional computation does not guarantee better selection: majority-vote accuracy can decline on tasks mixing easy and hard queries~\citep{chen2024more}, while verifier false positives can limit the accuracy achievable through resampling~\citep{stroebl2026limits}. These findings motivate improving the judgments used to select candidates. APT studies aspect-specific pairwise verification as one approach to this problem.

\textbf{LLM judges and verifiers.}
LLMs can evaluate responses through prompted judgments~\citep{zheng2023judging, liu2023geval, gu2024survey} or serve as trained generative verifiers for best-of-$n$ selection~\citep{zhang2025generative}. FLASK and Prometheus make evaluation more specific by decomposing quality into explicit criteria~\citep{ye2024flask, kim2024prometheus}; Constitutional AI similarly uses written principles to guide model-generated feedback~\citep{bai2022constitutional}. APT adopts criterion-based evaluation through separate aspect verifiers. Nevertheless, verification remains difficult. Self-verification and iterative refinement can improve responses~\citep{weng2023large, madaan2023selfrefine}, but intrinsic self-correction often fails on reasoning tasks~\citep{huang2024large}, and even strong judges struggle to distinguish objectively correct and incorrect responses on challenging pairs~\citep{tan2025judgebench}. Pairwise judgments are also sensitive to candidate order, particularly when responses are similar in quality~\citep{wang2024fair, shi2025judging}, and judges may favor their own generations~\citep{panickssery2024llm}. We address position bias in Section~\ref{sec:pairwise-verification} and discuss the implications of using same-family generators and verifiers in Section~\ref{sec:discussion}.

\textbf{Multi-agent evaluation and verification.}
Multi-agent LLM systems combine model outputs through interaction, as in debate~\citep{du2024improving, liang2024encouraging}, or through layered aggregation~\citep{wang2025mixture}. Debate has also been proposed for scalable oversight~\citep{irving2018ai}, with evidence that more capable debaters can help non-expert judges identify truthful answers~\citep{khan2024debating}. Other studies find that simple ensembling explains much of the improvement attributed to interaction: majority voting captures most of the gains from multi-agent debate~\citep{choi2025debate}, and sampling followed by voting improves with the number of agents~\citep{li2024more}. For evaluation, panels of smaller judges from different model families can outperform a single larger judge~\citep{verga2024replacing}, although verification remains a common failure point in multi-agent systems~\citep{cemri2025why}.

The closest verification framework is Multi-Agent Verification (MAV)~\citep{lifshitz2025multiagentverificationscalingtesttime}, which aggregates binary approvals from prompted aspect verifiers and demonstrates the value of scaling the verifier panel. MAV evaluates each candidate independently. APT instead asks each aspect verifier to compare two candidates, then combines their preferences by majority vote within each tournament match. We compare APT to BoN-MAV in Section~\ref{sec:experiments}.

\textbf{Pairwise comparison for test-time selection.}
Pairwise preferences have a long history in statistical ranking~\citep{bradley1952rank} and underpin preference-based model training~\citep{christiano2017deep, ouyang2022training} and large-scale model evaluation~\citep{chiang2024chatbot}. For LLM evaluation, pairwise judgments can agree more closely with human preferences than pointwise scores~\citep{zheng2023judging, liu2024aligning}. Pairwise prompting has also proved effective for zero-shot document ranking~\citep{qin2024large} and NLG evaluation~\citep{liusie2024llm}, while LLM-Blender trains a pairwise ranker to select among outputs from different models~\citep{jiang2023llmblender}.

Several works apply these ideas directly to test-time selection. PairJudge trains a pairwise reward model on PairJudge-432K and selects mathematical solutions through knockout tournaments~\citep{liu2025pairjudge}. \citet{guo2025reward} train reward reasoning models with reinforcement learning and aggregate pairwise judgments using Elo-based round-robin or knockout selection. \citet{zhou2025evaluating} evaluate LLM judges as selectors using round-robin comparisons in both candidate orders. On the theoretical side, \citet{chen2025provable} show that knockout selection can achieve vanishing failure probability under assumptions including a generator with nonzero success probability and a comparator that performs better than chance. Their repeated direct comparisons are closely related to our PV (no aspects) ablation in Section~\ref{sec:ablations}.

APT combines pairwise tournament selection with prompted multi-aspect verification. Its contribution is this training-free combination and its empirical evaluation, testing whether aspect-specific voting improves comparison over repeated generic judgments, and whether pairwise preferences provide useful selection information beyond binary approval (Section~\ref{sec:ablations}, Appendix~\ref{app:significance}).

\textbf{Tournament aggregation.}
Tournament rules determine how individual comparisons produce a final selection~\citep{moulin1986choosing, laslier1997tournament}. Single elimination requires only $n-1$ matches but depends on seeding~\citep{vu2009efficient}; Copeland's rule uses a full round-robin and provides a more comparison-intensive reference~\citep{copeland1951reasonable}. Work on noisy comparisons studies selection when individual judgments can be wrong~\citep{feige1994computing}, including settings where a comparison's error persists across repeated queries~\citep{braverman2008noisy}. Within each match, APT aggregates aspect-verifier votes. Condorcet's jury theorem motivates majority aggregation when voters are independent and individually better than chance~\citep{condorcet1785essai, grofman1983thirteen}, but shared errors among LLM verifiers limit its direct applicability. We empirically compare single-elimination, double-elimination, and Swiss-style tournaments in Appendix~\ref{app:tournament-formats}.

\section{Aspect-wise Pairwise Tournaments (APT)}
\label{sec:method}

We consider the standard best-of-$n$ setting. Given a question $q$, a generator LLM samples $n$ candidate solutions $\{o^{(1)}, \dots, o^{(n)}\}$, and a selection procedure returns one of them as the final output. APT selects through a single elimination tournament over the candidates. In each matchup, a set of aspect verifiers compares the two candidates, each along a different aspect such as logical soundness or domain knowledge, and each states which candidate it prefers. The candidate preferred by more verifiers advances, and the last candidate remaining is the final selection.

Section~\ref{sec:pairwise-verification} defines pairwise preference verification with a single judge. Section~\ref{sec:aspect-verifiers} decomposes each comparison across multiple aspect verifiers. Section~\ref{sec:tournament-aggregation} describes how the resulting matchup outcomes are combined into a single selection through a single elimination tournament, and Section~\ref{sec:APT} assembles the full algorithm.

\subsection{Pairwise Preference Verification}
\label{sec:pairwise-verification}

In pairwise preference verification, a judge is presented with two candidate solutions at once and asked which is better. Formally, given candidates $o^{(i)}$ and $o^{(j)}$, a judge $J$ produces
\begin{equation}
    \text{Pref}_J(o^{(i)}, o^{(j)}) \in \{i,\; j,\; \varnothing\},
\end{equation}
where $\varnothing$ denotes ``no preference.'' The ``no preference" option exists because the two options might both be equally correct or equally incorrect, and forcing a choice in those cases would add noise to the tournament while also possibly increasing token or thinking cost.


\textbf{Position bias.} LLM judges have a tendency to prefer candidates based on their given ordering in the prompt \citep{zheng2023judging, wang2024fair, shi2025judging}. We therefore query the judge twice per comparison, once with the candidates presented as $(o^{(i)}, o^{(j)})$ and once as $(o^{(j)}, o^{(i)})$, and keep the preference only if both orderings agree. When they disagree, the preference is assigned uniformly at random between the two candidates. We directly measure the position bias of our verifiers in Appendix~\ref{app:verifier-quality}.


\subsection{Decomposing Comparisons by Aspect}
\label{sec:aspect-verifiers}

\begin{figure}[t]
    \centering
    \includegraphics[width=0.8\linewidth]{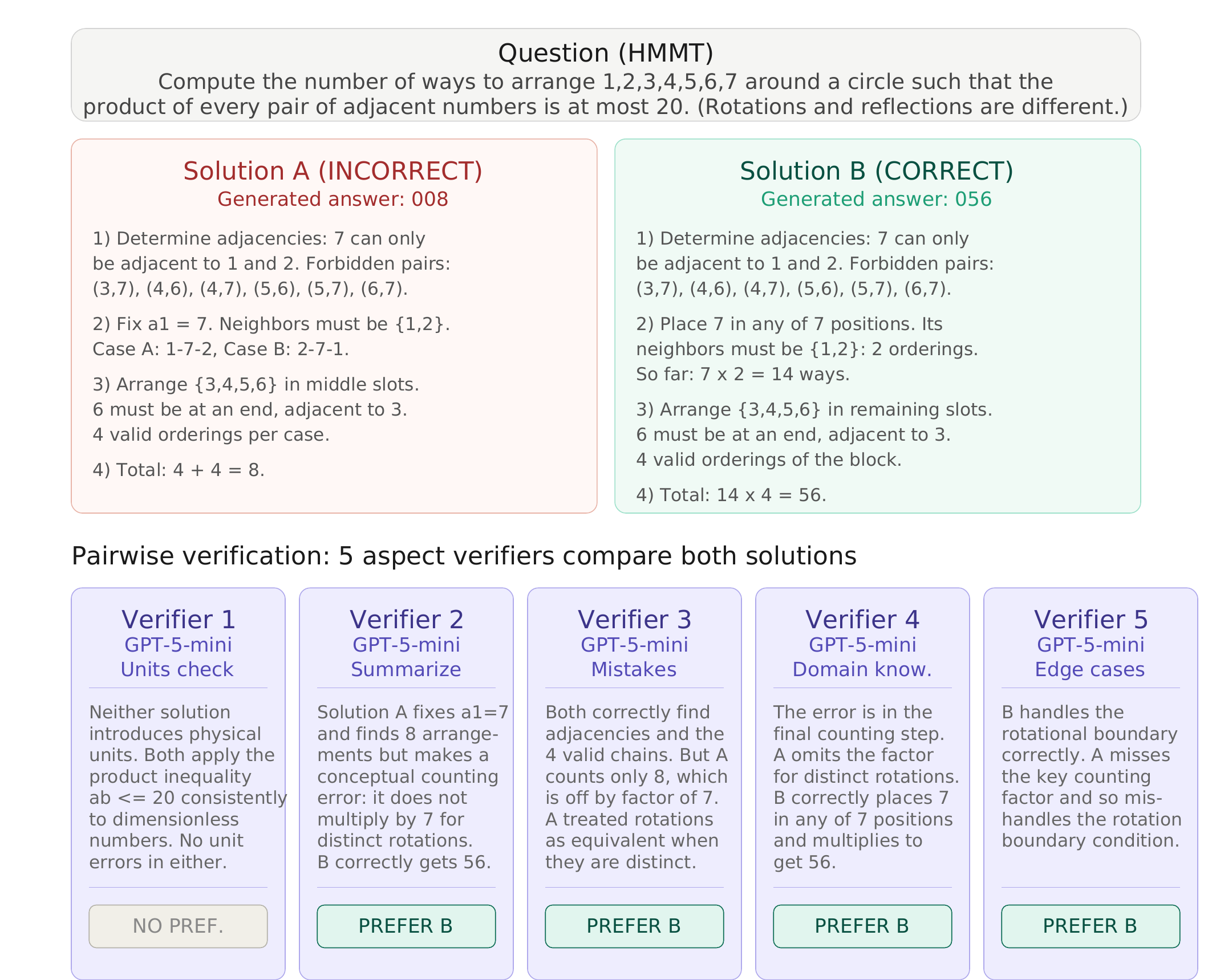}
    \caption{\textbf{Multi-aspect pairwise verification for a single matchup.} Two candidate solutions are compared by five aspect verifiers. Solution A finds 8 valid arrangements but fails to account for distinct rotations; Solution B correctly multiplies by 7 positions to get 56. Each verifier reads both full solutions and selects the one it judges to be better along its assigned aspect. Four of five verifiers identify Solution A’s counting error and prefer Solution B, while the units check verifier finds no meaningful difference. This illustrates the core advantage of pairwise verification: the rotation counting error in Solution A is subtle enough that it might be missed under independent binary evaluation, but becomes salient when a verifier can directly compare the two counting strategies side by side.}
    \label{fig:pairwise-verification}
    \vspace{-15pt}
\end{figure}

A single comparison asks the judge to weigh every dimension of quality at once, a task that gets harder with more complex candidates, so we instead decompose each comparison into several narrower ones. Following \citet{lifshitz2025multiagentverificationscalingtesttime}, we use aspect verifiers: off-the-shelf LLMs prompted to evaluate a specific property of a solution. Each aspect verifier is specified by a base LLM, the aspect it checks (e.g., logical soundness, domain knowledge), and a verification strategy (e.g., step-by-step analysis, identifying mistakes). In \citet{lifshitz2025multiagentverificationscalingtesttime}, each aspect verifier evaluates one candidate and returns a binary approval. In APT, each aspect verifier $v \in M$ instead acts as a pairwise judge restricted to its own aspect, producing $\text{Pref}_v(o^{(i)}, o^{(j)}) \in \{i, j, \varnothing\}$ (Figure~\ref{fig:pairwise-verification}).

The outcome of each matchup is decided by vote. Writing $c_i = \sum_{v \in M} \mathbb{1}[\text{Pref}_v(o^{(i)}, o^{(j)}) = i]$ for the number of verifiers preferring $o^{(i)}$, and $c_j$ analogously,
\begin{equation}
    \text{Winner}(o^{(i)}, o^{(j)}) =
    \begin{cases}
        i & \text{if } c_i > c_j \\[4pt]
        j & \text{if } c_j > c_i \\[4pt]
        \text{Uniform}\{i, j\} & \text{if } c_i = c_j
    \end{cases}
\end{equation}
Abstentions count toward neither side, so a matchup is decided by a plurality of the verifiers that express a preference rather than by a strict majority of $M$. Ties, including the case where every verifier abstains, are broken uniformly at random.

Section~\ref{sec:ablations} tests the contribution of aspects by comparing APT against a single generic judge.

\subsection{Tournament Aggregation}
\label{sec:tournament-aggregation}

\begin{figure}[t]
    \centering
    \includegraphics[width=0.8\linewidth]{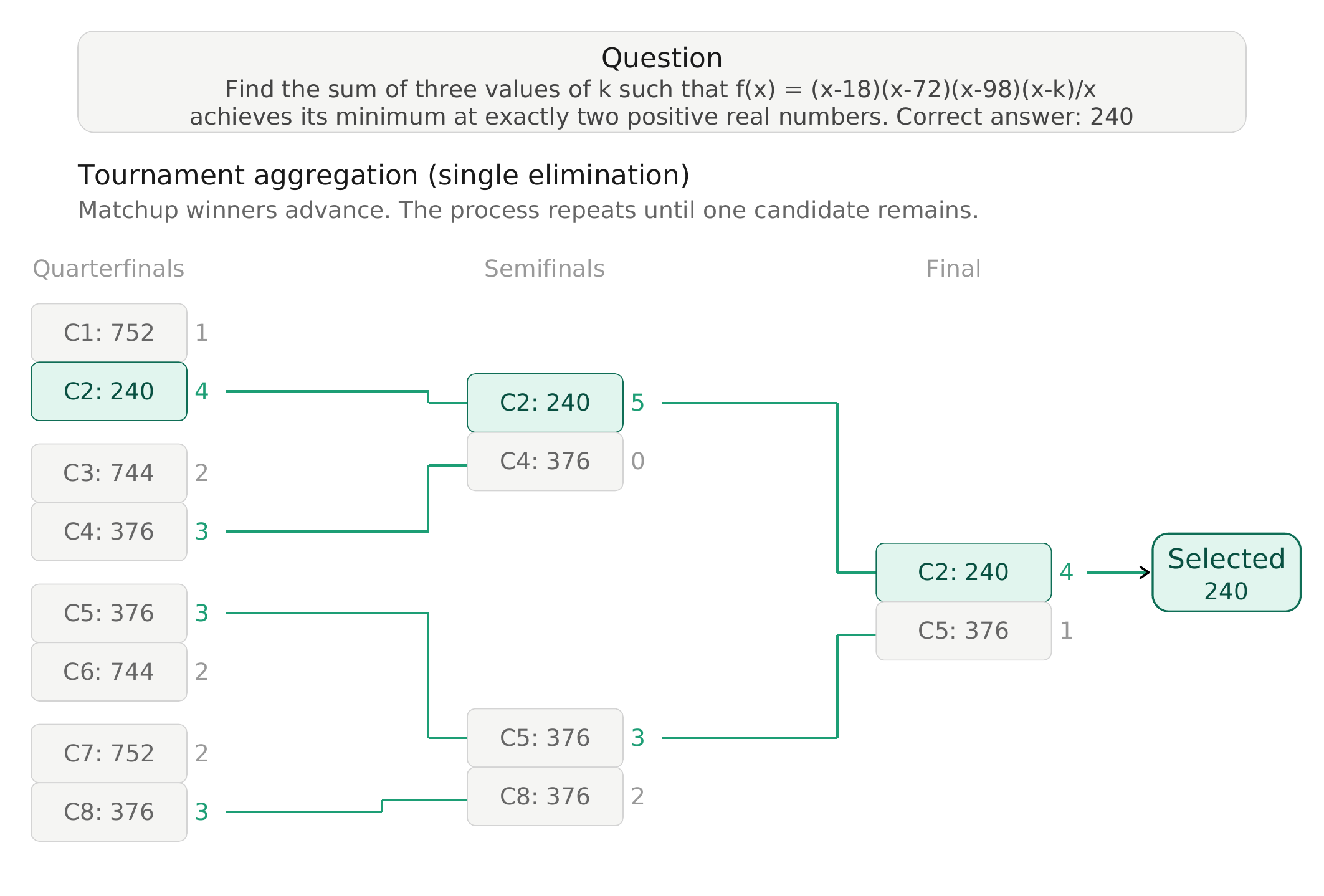}
    \vspace{-20pt}
    \caption{\textbf{Tournament aggregation for candidate selection.} Given 8 candidate solutions, a single elimination tournament selects the final output. At each matchup, aspect verifiers compare both solutions and the candidate preferred by the majority advances. The full tournament requires only $n - 1$ matchups, comparable in total verifier queries to binary approval methods which require $n$ independent evaluations.}
    \label{fig:tournament}
    \vspace{-10pt}
\end{figure}

Given $n$ candidate solutions sampled from a generator LLM, pairwise preferences must be aggregated into a single selected output. We adopt single elimination tournaments as the aggregation mechanism (Figure \ref{fig:tournament}). Candidates are paired, and the majority vote winner of each matchup advances to the next round. This process repeats until one candidate remains.

Single elimination requires exactly $n - 1$ pairwise matchups, making it the most compute-efficient tournament format. This efficiency does make it vulnerable to cases where a strong candidate could be eliminated early if it draws a difficult matchup. However, we find empirically that this is rarely a concern in practice (Appendix~\ref{app:bracket}).

\subsection{Aspect-wise Pairwise Tournaments}
\label{sec:APT}


We introduce Aspect-wise Pairwise Tournaments (APT) as a test-time compute algorithm that uses pairwise preference tournaments to aggregate judgements from aspect verifiers. APT consists of three steps:
\begin{enumerate}
    \item \textbf{Generate:} Sample $n$ candidate solutions from a generator LLM.
    \item \textbf{Compare:} For each matchup in the single elimination bracket, present both candidate solutions to a set of $m$ aspect verifiers and determine the winner by majority vote over the verifiers' preferences.
    \item \textbf{Select:} Return the tournament winner as the final output.
\end{enumerate}

The APT algorithm is detailed in Algorithm \ref{alg:APT}. Formally, the selected output is:
\begin{equation}
    \hat{o} = \text{SingleElim}(\{o^{(1)}, \dots, o^{(n)}\}, M)
\end{equation}
where the $\text{SingleElim}$ function applies a single elimination tournament using pairwise matchups judged by the verifier set $M$. 

APT scales test-time compute along two dimensions: the number of candidate solutions $n$ and the number of verifiers $m$. Increasing $n$ provides more candidate diversity, while increasing $m$ improves the reliability of each pairwise comparison. We test these in Appendix~\ref{app:scaling} and Appendix~\ref{app:verifier-ablation}. Generation can be fully parallelized since all candidates are sampled independently. Each round of the single elimination tournament can also be parallelized, as the each of the verifiers and matchups are independent from the others.

\section{Experiments}
\label{sec:experiments}

We evaluate APT across five domains using three generator LLMs. We aim to address the following questions: (1) how APT compares to existing methods; (2) whether it can be steered toward user preferences; and (3) what each component, aspects and pairwise preferences, contributes.

\textbf{Domains.}
We evaluate on five domains spanning mathematical reasoning, general knowledge, and strategic reasoning:

\begin{itemize}
    \item \textbf{AIME.} Competition mathematics problems from the American Invitational Mathematics Examination (2022-2026 exams), 150 problems total
    
    \item \textbf{HMMT.} Problems from the Harvard-MIT Mathematics Tournament (February and November 2025, February 2026), 93 problems total.
    
    \item \textbf{HLE.} A subset of the Humanity's Last Exam (HLE) benchmark \citep{phan2025hle}, filtered to text-only multiple-choice questions for ease of evaluation. Problems span diverse academic domains (biology, medicine, science, etc.). Tested on 300 problems. 
    
    \item \textbf{Chess Validity.} A state-tracking task over chess games. Given a sequence of moves with the final move's destination square omitted, the model must supply a valid destination square. Problems were drawn from synthetic games with short length. Tested on 300 problems. 
    
    \item \textbf{Chess Optimality.} A strategic reasoning task over positions from grandmaster games. Given the first 13 moves of a game, the model must suggest the best 14th move for White. Unlike the other datasets, this uses continuous scoring via Stockfish (depth 20), with the score reported in pawn units. Tested on 300 problems. 
\end{itemize}

\textbf{Generator and Verifier Setups.}
We evaluate three generator models spanning different model families and capability levels: Gemini-3.1-Pro, GPT-5.2, and Qwen3-8B. For each model, we sample $n = 8$ candidate solutions per problem. Generation with Gemini-3.1-Pro used Gemini-2.5-Flash as the verifier model while generation with GPT-5.2 and Qwen3-8B used GPT-5-mini as the verifier model. Appendix \ref{app:experimental-setup} includes details on verifier setups and generator/verifier prompts. 

\textbf{Baselines.}
We compare APT against the following test-time compute methods, which all also use $n = 8$ candidate solutions:
\begin{itemize}
    \item \textbf{Self-Consistency}: Most common answer among 8 candidate solutions \citep{wang2023selfconsistency}. Ties broken arbitrarily. 
    \item \textbf{BoN-RM}: Answer with highest reward model score, using Skywork-Reward-V2-Llama-3.1-8B \citep{liu2025skywork}, which is currently the top performing model on RewardBench2 \citep{RewardBench2}.
    \item \textbf{BoN-MAV}: Answer with highest number of approvals from binary output aspect verifiers \citep{lifshitz2025multiagentverificationscalingtesttime}. Ties are broken uniformly at random among the tied candidates, since we take equal approval counts to indicate equal estimated quality. BoN-MAV uses the same aspect verifiers and the same base verifier models as APT.
\end{itemize}

\subsection{Main Results}
\label{sec:main-results}

Table~\ref{tab:main-results} presents the main results across all three generator models and five domains. We report accuracy (\%) for AIME, HMMT, HLE, and Chess Validity, and pawn score for Chess Optimality (higher is better for all metrics). APT is compared against self-consistency, reward model verification, and binary MAV, all using $n = 8$ candidate solutions.

\begin{table}[t]
\centering
\small
\begin{tabular}{llccccc}
\toprule
\textbf{Generator LLM} & \textbf{Method} & \textbf{AIME} & \textbf{HMMT} & \textbf{HLE} & \textbf{Chess Val.} & \textbf{Chess Opt.} \\
\midrule
\multirow{4}{*}{Gemini-3.1-Pro}
 & Self-Consistency  & 96.67 & 50.54 & 48.00 & \textbf{99.00} & 0.031 \\
 & BoN-RM            & 97.33 & 50.54 & 43.33 & 98.00 & 0.107 \\
 & BoN-MAV           & 96.67 & 50.54 & 46.33 & 97.33 & 0.056 \\
 & APT      & \textbf{98.00} & \textbf{54.84} & \textbf{49.67}$^\dagger$ & 98.67$^\dagger$ & \textbf{0.166}$^\dagger$ \\
\midrule
\multirow{4}{*}{GPT-5.2}
 & Self-Consistency  & 66.00 & 26.88 & 12.00 & 52.67 & 1.866 \\
 & BoN-RM            & 68.00 & 30.11 & 11.67 & 58.33 & 1.666 \\
 & BoN-MAV           & 78.00 & 36.56 & 12.67 & 62.00 & 1.856 \\
 & APT      & \textbf{82.67}$^\dagger$ & \textbf{39.78} & \textbf{15.00}$^\dagger$ & \textbf{70.00}$^\dagger$ & \textbf{1.982}$^\dagger$ \\
\midrule
\multirow{4}{*}{Qwen3-8B}
 & Self-Consistency  & 21.33 & 7.53  & 11.33 & 10.67 & 0.586 \\
 & BoN-RM            & 24.00 & 6.45  & 12.00 & 11.00 & 0.577 \\
 & BoN-MAV           & 28.67 & 7.53  & 11.33 & 10.33 & 0.677 \\
 & APT      & \textbf{31.33} & \textbf{10.75} & \textbf{14.33} & \textbf{15.67} & \textbf{1.221} \\
\bottomrule
\end{tabular}
\caption{\textbf{Main results across models and domains.} Performance comparison of four test-time verification methods using $n = 8$ candidate solutions across three generator LLMs. Bold indicates the best result among methods within each model group. Accuracy (\%) is reported for AIME, HMMT, HLE, and Chess Validity; pawn score is reported for Chess Optimality. Higher is better. $^\dagger$APT exceeds BoN-MAV with a 95\% paired-bootstrap interval that excludes zero (Table~\ref{tab:significance}, Appendix~\ref{app:significance}); Qwen3-8B was not tested.}
\label{tab:main-results}
\vspace{-15pt}
\end{table}

\textbf{APT outperforms all other baselines in nearly all model-domain pairs.} APT achieves the highest score in 14 out of 15 model-domain pairs. We find that APT yields the largest gains in performance on the open-ended task, Chess Optimality. On these problems, since there is no single correct answer, the combination of multiple aspects and pairwise comparisons force the verifier to make a more well-rounded judgement. These types of judgements are where we would expect the method to transfer (Section~\ref{sec:discussion}). For a similar reason, we would expect APT to improve on Humanity's Last Exam well, since the difficulty of the problems themselves also lead to more difficult verification tasks, where the combination of multiple aspects and pairwise comparison could be more illustrative for verification. However, we also observe that the gains from APT on the HLE dataset is limited by the ability of the generator to create valid solutions (Appendix~\ref{app:failure}). Improvements from APT are also more limited in already saturated domains, like Gemini-3.1-Pro on AIME or Chess Validity; here the marginal benefit of APT is small because there is limited room to improve. We report additional failure analysis results in Appendix~\ref{app:failure}.



\textbf{Weak-to-strong generalization.} We also observe that APT enables weak-to-strong generalization, i.e. combining weaker verifiers can improve the outputs of stronger generators. In our setup, aspect verifiers use smaller models from the same family as the generator (GPT-5-mini verifiers for GPT-5.2 generation, Gemini-2.5-Flash verifiers for Gemini-3.1-Pro generation). Despite this capability gap, APT consistently improves solution quality over both self-consistency and reward model baselines. This suggests that even when a weaker verifier may struggle to determine whether a solution from a stronger generator is correct in isolation, it can still reliably identify which of two such solutions is better. This further justifies our use of pairwise comparisons, as these judgments can be made using weaker models, enabling APT to be carried out even on small compute budgets. 

\textbf{Self-improvement} We also investigate whether APT can enable self-improvement, where the same base model serves as both the generator and the set of aspect verifiers. We evaluate this setting using two models: Gemini-2.5-Flash and GPT-5-mini, which are both relatively small and cost-effective compared to their larger analogues used in the main experiments.
 
As shown in Table~\ref{tab:self-improvement}, APT with self-verification consistently achieves the best performance across both models in all evaluated settings. Across both models, pairwise self-verification provides the strongest selection signal on the harder and more diverse benchmarks (such as HMMT, HLE, and Chess Optimality), while self-consistency remains competitive on tasks where accuracy is already high (such as AIME). APT with self-verification is also more accurate than BoN-MAV with self verification, indicating that pairwise preferences might set an easier task for smaller models. These results demonstrate that APT does not require an exceptionally capable generator model to be effective. Even smaller models can improve their own outputs by comparing candidate solutions head-to-head.

\begin{table}[t]
\centering
\small
\begin{tabular}{llccccc}
\toprule
\textbf{Model} & \textbf{Method} & \textbf{AIME} & \textbf{HMMT} & \textbf{HLE} & \textbf{Chess Val.} & \textbf{Chess Opt.} \\
\midrule
\multirow{4}{*}{Gemini-2.5-Flash}
 & Self-Consistency  & 88.00 & 37.63 & 8.00 & 62.67 & 0.035 \\
 & BoN-MAV           & 84.67 & 32.26 & 10.33 & 62.00 & 0.047 \\
 & APT      & \textbf{90.00} & \textbf{38.71} & \textbf{13.33} & \textbf{73.00} & \textbf{0.086} \\
\midrule
\multirow{4}{*}{GPT-5-mini}
 & Self-Consistency  & 53.33 & 25.81 & 10.00 & 43.67 & 0.611 \\
 & BoN-MAV           & 57.33 & 29.03 & 10.67 & 49.00 & 0.614\\
 & APT      & \textbf{64.67} & \textbf{35.48} & \textbf{11.33} & \textbf{53.33} & \textbf{0.695} \\
\bottomrule
\end{tabular}
\caption{\textbf{Self-improvement experiments.} The same (small) model is used as both generator and verifier.}
\label{tab:self-improvement}
\vspace{-15pt}
\end{table}

\subsection{Steering Selection Toward User Preferences}
\label{sec:steerability}

Because each aspect verifier is simply a prompt, APT lets users decide what ``better'' means. A user who cares about a property beyond correctness can add an aspect for it, and weight that aspect's vote according to how much it matters to them. This lets APT select outputs that are aligned with a user's preferences, without any training.

\textbf{Setup.} We add one new aspect verifier to the existing set: \emph{conciseness} on AIME and HMMT, and \emph{aggressiveness} (a preference for attacking, initiative-seeking play) on Chess Optimality. Both use the same base verifier models as our main experiments. Each matchup is decided by a weighted vote, $c_i = \sum_{v \in M} w_v \, \mathbb{1}[\text{Pref}_v(o^{(i)}, o^{(j)}) = i]$, where existing verifiers keep $w_v = 1$ and the new aspect receives weight $w \in \{0, 1, 2, 3\}$, so $w = 0$ recovers standard APT. We reuse the $n = 8$ candidates from our main experiments. For conciseness we measure the token length of the selected solution. For aggressiveness we measure whether the selected move is forcing (a check or a capture), determined by parsing the move against the board position.

\begin{figure}[t]
    \centering
    \includegraphics[width=\linewidth]{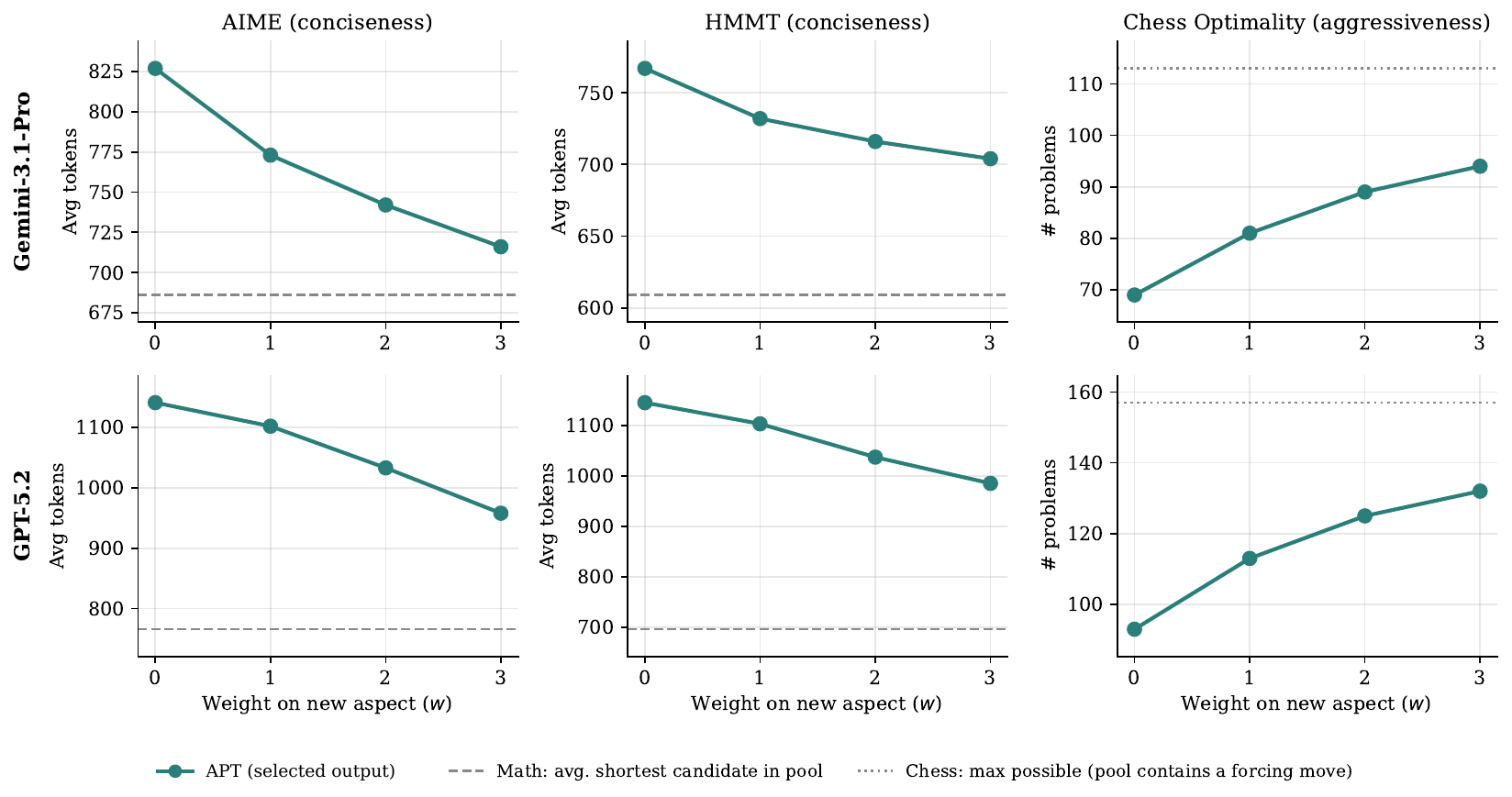}
    \caption{\textbf{Steering APT toward user-specified preferences.} A single new aspect verifier is added to the existing set and its vote is weighted by $w$. \textbf{Left, Center:} a conciseness aspect on AIME and HMMT, measured by the average token length of the selected solution. The dashed line is the average length of the shortest candidate in each pool, the lowest achievable value. \textbf{Right:} an aggressiveness aspect on Chess Optimality, measured by the number of problems where the selected move is forcing (a check or capture). The dotted line is the number of problems whose candidate pool contains at least one forcing move.}
    \label{fig:steerability}
    \vspace{-15pt}
\end{figure}

\textbf{Results.} Upweighting the new aspect moves the selected output toward the user's preference in every setting (Figure~\ref{fig:steerability}). Going from $w = 0$ to $w = 3$, the average length of the selected solution decreases by 13\% (Gemini-3.1-Pro) and 16\% (GPT-5.2) on AIME, and by 8\% and 14\% on HMMT. On Chess Optimality, the number of problems where APT selects a forcing move increases by 36\% (Gemini-3.1-Pro) and 42\% (GPT-5.2). Measured against the best achievable value in each pool, the most concise candidate or the maximum number of available forcing moves, $w = 3$ closes 36--79\% of the gap on the math tasks and 57--61\% on chess.

Importantly, the new aspect adds a preference on top of the existing verifiers rather than replacing them. The existing verifiers continue to vote on correctness, so APT is inclined to select the most concise solution \emph{out of the existing set of correct answers}. We saw that accuracy on AIME and HMMT decreases by at most three problems in any generator-dataset-weight setting, which shows steering toward conciseness comes at little cost to correctness.

\subsection{Ablations and Additional Results}
\label{sec:ablations}

We first check that the main results do not depend on the choice of $n$. We then test the two components of APT separately. The comparison against BoN-MAV in Table~\ref{tab:main-results} already isolates the pairwise structure, since BoN-MAV uses the same aspect verifiers and base models; the ablations below isolate the aspect vote, by replacing it with a generic judge and by varying the number of aspect verifiers. 

\textbf{Scaling the number of candidates.} Table~\ref{tab:main-results} fixes $n = 8$, leaving open whether the ordering of methods is an artifact of that choice. We re-ran every method at every $n \in \{1, \dots, 16\}$ for both closed-source generators and all five datasets (Figure~\ref{fig:scaling}, Appendix~\ref{app:scaling}). All three methods improve with more candidates, though not at the same rate. In every generator-dataset combination except Gemini-3.1-Pro on Chess Validity and GPT-5.2 on Chess Optimality, APT@$n$ is at or above both baselines over the full range of $n$, and in several cases (Gemini-3.1-Pro on HLE, GPT-5.2 on Chess Validity), the gap widens with $n$ rather than closing. The $n = 8$ results therefore understate the method's full potential.

\textbf{Aspect voting versus a single generic judge.} We construct PV-1, identical to APT with the same tournament setup, the same position-bias handling, and the same base verifier model, except that each matchup is decided by a direct ``which solution is better?'' query with no aspect-specific instruction. This is the closest analogue in our setup to the comparator of \citet{chen2025provable}. One such query per matchup uses fewer verifier calls than APT's $m = 5$ verifiers, so we also report a budget-matched variant in which five independent generic judgments are sampled per matchup and aggregated by the same majority rule.

\begin{table}[t]
\centering
\small
\begin{tabular}{lccc|ccc}
\toprule
& \multicolumn{3}{c|}{\textbf{Gemini-3.1-Pro}} & \multicolumn{3}{c}{\textbf{GPT-5.2}} \\
\textbf{Dataset} & \textbf{APT} & \textbf{PV-1} & \textbf{PV-5} & \textbf{APT} & \textbf{PV-1} & \textbf{PV-5} \\
\midrule
AIME        & \textbf{98.00} & 94.67 & 95.33 & \textbf{82.67} & 66.67 & 69.33 \\
HMMT        & \textbf{54.84} & 47.31 & 49.46 & \textbf{39.78} & 27.96 & 31.18 \\
HLE         & \textbf{49.67} & 42.67 & 43.33 & \textbf{15.00} & 10.33 & 11.00 \\
Chess Val.  & \textbf{98.67} & 96.00 & 96.67 & \textbf{70.00} & 59.33 & 61.67 \\
Chess Opt.  & \textbf{0.166} & 0.051 & 0.093 & \textbf{1.982} & 1.314 & 1.519 \\
\bottomrule
\end{tabular}
\caption{\textbf{The aspect vote is not cosmetic.} APT compared against pairwise tournaments whose matchups are decided by a generic ``which solution is better?'' query instead of by a vote over aspect verifiers, at $n = 8$. PV-1 samples one generic judgment per matchup; PV-5 samples five and aggregates them by the same majority rule as APT, matching APT's verification budget. Verifiers are Gemini-2.5-Flash and GPT-5-mini respectively. Accuracy (\%) except Chess Optimality (pawns).}
\label{tab:ablation-aspects}
\vspace{-15pt}
\end{table}

APT outperforms both variants on every generator-dataset pair (Table~\ref{tab:ablation-aspects}). Most of what the aspect vote contributes therefore comes from the diversity of the criteria rather than from the extra samples. Also, five draws from one generic judge are susceptible to correlated errors while five verifiers along different axes can diversity errors. Thus, we find that both the pairwise structure without aspects (PV-5) and the aspects without the pairwise structure (BoN-MAV) are weaker than APT.

\textbf{Number of aspect verifiers.} Removing verifiers from the full set one at a time and recomputing accuracy for each resulting subset (Figure~\ref{fig:verifier-ablation}, Appendix~\ref{app:verifier-ablation}) shows accuracy increasing with the number of verifiers for both APT and BoN-MAV on all five datasets, so the aspects contribute complementary rather than redundant signal, and performance does not rely on any single aspect. APT is at or above BoN-MAV at nearly every verifier count, so the pairwise advantage is not an artifact of a particular number or subset of aspects.

\textbf{Robustness and signal quality.} Three further measurements, reported in full in the appendix, support the design. Sampling five random bracket seedings per problem changes APT's score by little relative to the gaps in Table~\ref{tab:significance} (Appendix~\ref{app:bracket}), because most problems admit several correct candidates and a poor draw for one does not affect the rest. The verifier majority prefers the correct solution in 78.4-92.3\% of correct-versus-incorrect matchups across domains and models (Appendix~\ref{app:verifier-quality}), and individual verifier preferences are transitive in 89-98\% of candidate triples. These explain why the aggregation rule barely matters (Appendix~\ref{app:tournament-formats}) and provides the basis for the use of single-elimination tournaments. 

\section{Discussion and Future Work}
\label{sec:discussion}

We have introduced APT, which verifies candidate solutions at test-time in the same way people tend to choose between complex options, by comparing them one aspect at a time. Each matchup of a knockout tournament is decided by a vote across off-the-shelf aspect verifiers, with each verifier stating a pairwise preference along its own aspect. Pairwise comparison and multi-aspect verification each have precedent, but our ablations show the combination is not redundant, as binary approval over the same aspects and a generic pairwise judge at a matched budget both fall short of APT. The method also enables weak-to-strong generalization and self-improvement in weaker models, and requires no training at any stage, so adapting it to a new domain is easy.

We find the gains from APT are not uniform. They are largest on Chess Optimality, the one open-ended task in our suite, where no strict answer key exists, and a verifier has more leeway in determining which candidates are ``better". They are smallest on near-saturated benchmarks, where most candidates are already correct and majority voting is hard to beat. This suggests that pairwise, aspect-level comparison matters most when quality is measured in degrees rather than a binary property that can be checked in isolation.

Decomposing verification into aspects has a second benefit beyond accuracy. It also makes the notion of ``better'' explicit and editable. Adding a single weighted aspect was enough to steer selection toward more concise math solutions and more aggressive chess moves, with little cost to correctness on the math tasks, and without needing to generate new candidates. A trained judge can only encode a fixed notion of quality, but APT's can be changed by the user at inference time. Natural extensions include steering along several preferences at once, generating aspects automatically from a user's description of what they want, and applying steering in domains where preferences vary more widely across users.

\paragraph{Limitations.}
We only test generators and verifiers from the same model family. The Qwen3-8B setting, which uses GPT-5-mini verifiers, offers a partial cross-family check, but swapping verifier families would test this directly. Most of our evaluation is also on domains with verifiable answers. We did not evaluate summarization, long-form writing, or code generation, though the Chess Optimality results suggest these are where both the accuracy and steering benefits of APT are most likely to matter. Finally, although APT requires no training, the tournament adds inference cost. With $n$ candidates, $k$ aspect verifiers, problem-prompt length $p$, and mean solution length $s$, APT issues $2(n-1)k$ verifications over the problem plus two solutions, since each comparison is queried in both orderings, costing roughly $2(n-1)k(p+2s)$ input tokens. APT is therefore up to about four times as expensive as BoN-MAV in input tokens when $s \gg p$. It is also slower: BoN-MAV's verifications run in a single parallel batch, whereas tournaments require sequential rounds.

\bibliography{main}
\bibliographystyle{tmlr}

\newpage

\appendix
\section{Appendix}

\subsection{APT Algorithm}

\begin{algorithm}[h]
\caption{APT: Aspect-wise Pairwise Tournaments}
\label{alg:APT}
\KwIn{Question $q$, generator LLM $G$, aspect verifiers $M = \{v_1, \dots, v_m\}$, number of candidates $n$}
\KwOut{Selected solution $\hat{o}$}

\textbf{Step 1: Generate candidates} \\
Sample $n$ candidate solutions $\{o^{(1)}, \dots, o^{(n)}\}$ from $G$ given $q$ \\[6pt]

\textbf{Step 2: Single elimination tournament} \\
$\mathcal{S} \leftarrow \{o^{(1)}, \dots, o^{(n)}\}$ \tcp*{active candidates}
\While{$|\mathcal{S}| > 1$}{
    Pair candidates in $\mathcal{S}$ into matchups \\
    $\mathcal{S}' \leftarrow \emptyset$ \\
    \ForEach{matchup $(o^{(i)}, o^{(j)})$}{
        \tcp{Pairwise preference voting}
        $\text{votes}_i \leftarrow 0, \quad \text{votes}_j \leftarrow 0$ \\
        \ForEach{verifier $v_k \in M$}{
            $p \leftarrow \text{Pref}_{v_k}(o^{(i)}, o^{(j)})$ \tcp*{query in both orderings; see \S\ref{sec:pairwise-verification}}
            \lIf{$p = i$}{$\text{votes}_i \leftarrow \text{votes}_i + 1$}
            \lElseIf{$p = j$}{$\text{votes}_j \leftarrow \text{votes}_j + 1$}
            \lElse{abstain \tcp*{no preference}}
        }
        \uIf{$\text{votes}_i > \text{votes}_j$}{$\mathcal{S}' \leftarrow \mathcal{S}' \cup \{o^{(i)}\}$}
        \uElseIf{$\text{votes}_j > \text{votes}_i$}{$\mathcal{S}' \leftarrow \mathcal{S}' \cup \{o^{(j)}\}$}
        \Else{$\mathcal{S}' \leftarrow \mathcal{S}' \cup \{\text{Uniform}\{o^{(i)}, o^{(j)}\}\}$ \tcp*{tie or all abstain}}
    }
    $\mathcal{S} \leftarrow \mathcal{S}'$ \\
}

\textbf{Step 3: Return winner} \\
$\hat{o} \leftarrow$ the single remaining element of $\mathcal{S}$ \\
\Return{$\hat{o}$}
\end{algorithm}

\subsection{Experimental Setup} \label{app:experimental-setup}

Table \ref{tab:verifiers} lists the aspect verifiers used for each of the five domains tested. 

\begin{table*}[h]
\centering
\small
\begin{tabular}{lllcccc}
\toprule
\textbf{Base Model} & \textbf{Aspect} & \textbf{Strategy} & \textbf{\shortstack{AIME/\\HMMT}} & \textbf{HLE} & \textbf{\shortstack{Chess\\Val.}} & \textbf{\shortstack{Chess\\Opt.}} \\
\midrule
\multirow{8}{*}{\shortstack[l]{GPT-5-mini /\\Gemini-2.5-Flash}}
 & Unit Conversions    & Step-by-Step      & $\bullet$ &           &           &           \\
 & General Correctness & Summarize         & $\bullet$ &           & $\bullet$ & $\bullet$ \\
 & General Correctness & Identify Mistakes & $\bullet$ & $\bullet$ & $\bullet$ & $\bullet$ \\
 & Domain Knowledge    & Direct            & $\bullet$ & $\bullet$ & $\bullet$ & $\bullet$ \\
 & Edge Cases          & Direct            & $\bullet$ &           &           &           \\
 & Logical Soundness   & Step-by-Step      &           & $\bullet$ & $\bullet$ & $\bullet$ \\
 & General Correctness & Diff Comparison   &           & $\bullet$ &           &           \\
 & Factual Accuracy    & Step-by-Step      &           & $\bullet$ & $\bullet$ & $\bullet$ \\
\bottomrule
\end{tabular}
\caption{\textbf{Aspect verifiers used per domain.} Each verifier is defined by a base model and an aspect-strategy pair. The specific base model depends on the generator model family (i.e. Gemini verifiers used for Gemini generations). Verifiers marked with $\bullet$ are included in the domain-specific subset for that domain.}
\label{tab:verifiers}
\end{table*}

The generator and verifier models used throughout were: \texttt{gemini-3.1-pro-preview} \citep{google2026gemini31pro}, \texttt{gemini-2.5-flash} \citep{google2025gemini25flash}, \texttt{gpt-5.2-2025-12-11} \citep{openai2025gpt52}, \texttt{gpt-5-mini-2025-08-07} \citep{openai2025gpt5}, and \texttt{Qwen3-8B} \citep{yang2025qwen3}. 

We used domain-specific prompts for the generator LLMs, which are documented in Table \ref{tab:gen-prompts}. Prompts for aspect verifiers consisted of domain-specific system prompts combined with an aspect-specific verification prompt. These are documented in Tables \ref{tab:vera-system-prompts}
 and \ref{tab:vera-aspect-prompts}. 

\paragraph{Chess Optimality Scoring.} A proposed move is scored by playing it on the board and evaluating the resulting position with Stockfish at search depth 20. The score is the evaluation in pawn units (centipawns divided by 100) from White's perspective, so higher values indicate better moves. Positions with a forced mate are assigned ±100. Illegal or unparseable moves, and responses from which no move can be extracted, receive a neutral score of 0.

\subsection{Additional Qualitative Illustrations}
\label{app:illustrations}

\begin{figure}[t]
    \centering
    \includegraphics[width=0.9\linewidth]{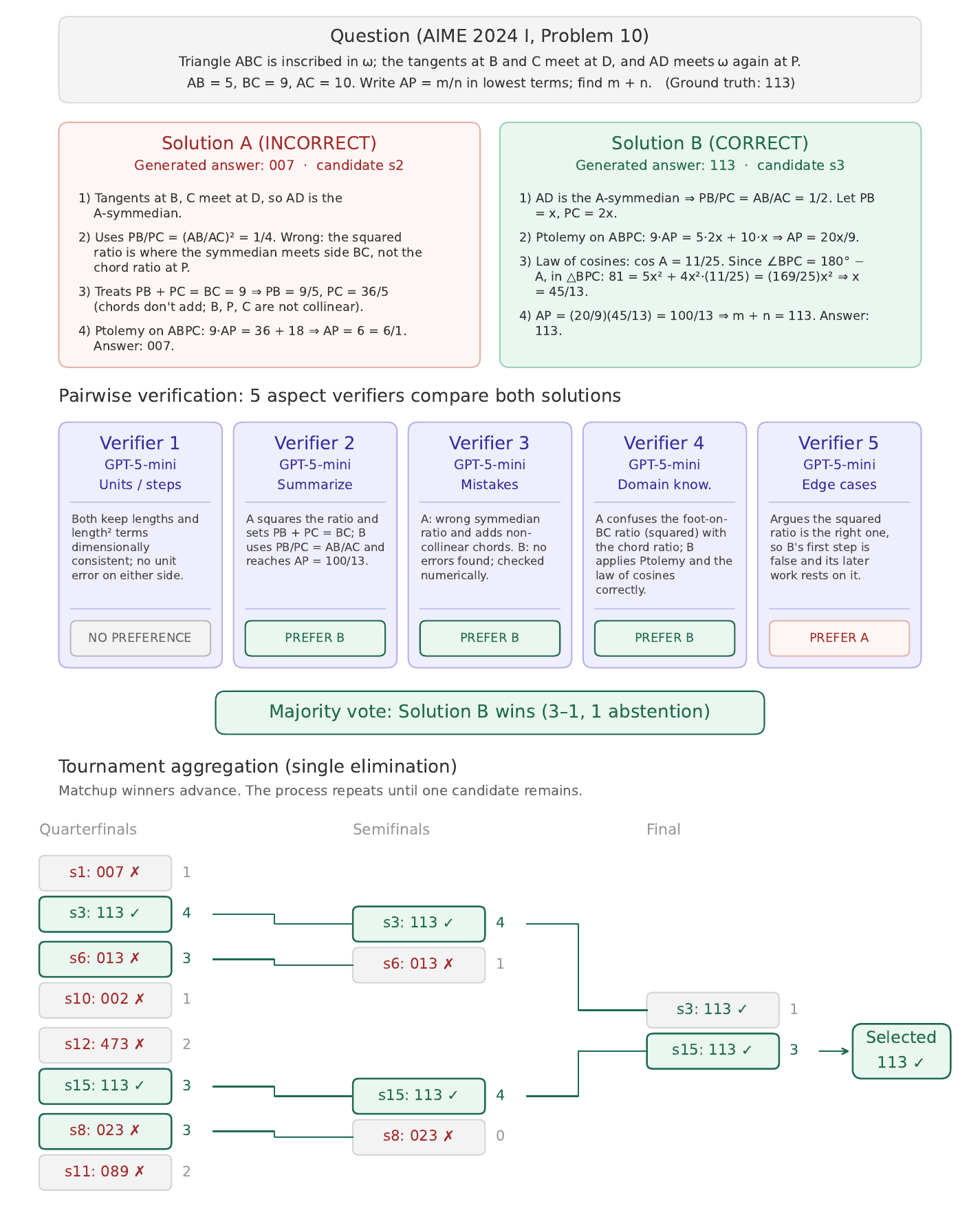}
    \caption{\textbf{APT applied to AIME problem.} Top: Pairwise verification for a matchup between an incorrect solution and a correct solution. Both solutions identify $AD$ as the $A$-symmedian, but Solution A uses the wrong ratio for $PB/PC$, squaring $AB/AC$, and then wrongly assumes $PB + PC = BC$. Three of five verifiers detect these errors and prefer Solution B; the units verifier abstains, and the edge-case verifier is misled into accepting A's squared ratio. This illustrates how majority vote across diverse aspects provides robustness to individual verifier mistakes. Bottom: The full single elimination tournament over 8 candidates (2 correct, 6 incorrect). Incorrect candidates can advance past other incorrect candidates, but the correct solutions win every matchup against incorrect ones; both reach the final, and one is selected as the output.}
    \label{fig:aime-illustration}
\end{figure}

\begin{figure}[t]
    \centering
    \includegraphics[width=0.85\linewidth]{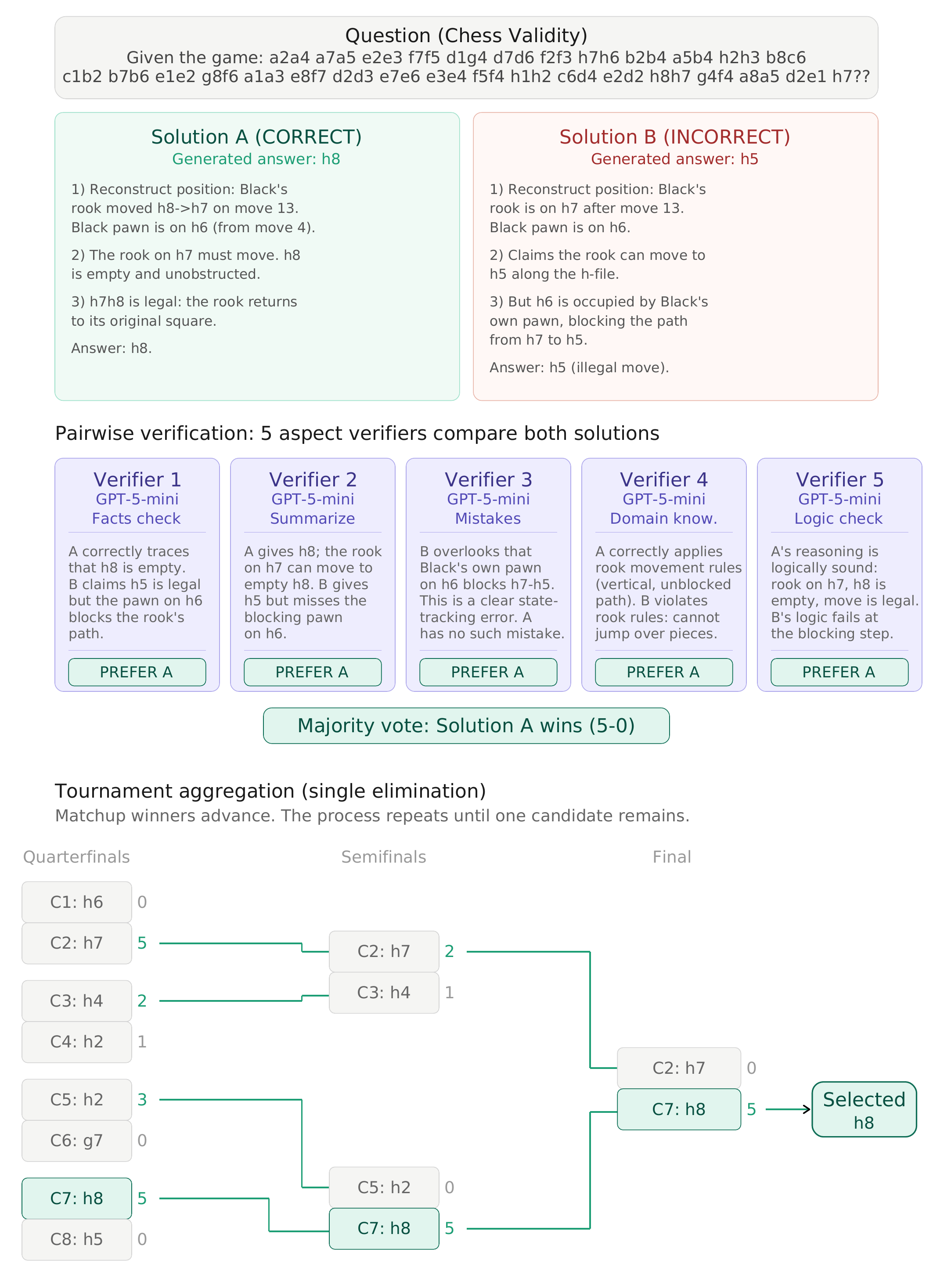}
    \caption{\textbf{APT applied to a chess move validity problem.} Top: Pairwise verification for a quarterfinal matchup. Given a partial game, the model must supply the destination square for the final move. Solution A correctly identifies that the rook on h7 can return to the empty h8 square; Solution B proposes h5 but overlooks that Black's own pawn on h6 blocks the rook's path. All five verifiers unanimously prefer Solution A. Bottom: The full tournament over 8 candidates, of which only one is correct. Despite 7 of 8 candidates being incorrect, pairwise verification reliably eliminates them.}
    \label{fig:chess-illustration}
\end{figure}

Figures \ref{fig:aime-illustration} and \ref{fig:chess-illustration} provide additional examples of how APT selects a final solution based on a set of candidate solutions. Figure \ref{fig:aime-illustration} illustrates APT on a problem from the 2024 AIME I exam. Figure \ref{fig:chess-illustration} demonstrates the algorithm on a chess move validity problem. Both figures are designed in the same style as Figures \ref{fig:pairwise-verification} and \ref{fig:tournament}.

\subsection{Testing Significance}
\label{app:significance}

\begin{table}[h]
\centering
\small
\begin{tabular}{lcc}
\toprule
\textbf{Comparison} & \textbf{Gemini-3.1-Pro} & \textbf{GPT-5.2} \\
\midrule
AIME $-$ cons            & $+1.33$ [$-0.67$, $+3.33$] & $\mathbf{+16.67}$ [$+10.67$, $+22.67$] \\
AIME $-$ BoN-MAV         & $+1.33$ [$-0.67$ , $+4.00$] & $\mathbf{+4.67}$ [$+1.33$, $+10.67$] \\
HMMT $-$ cons            & $\mathbf{+4.30}$ [$+1.08$, $+8.61$] & $\mathbf{+12.90}$ [$+4.30$, $+21.51$] \\
HMMT $-$ BoN-MAV         & $+4.30$ [$+0.00$, $+9.68$] & $+3.23$ [$-1.08$, $+8.60$] \\
HLE $-$ cons             & $\mathbf{+1.67}$ [$+0.33$, $+4.33$] & $\mathbf{+3.00}$ [$+1.00$, $+5.33$] \\
HLE $-$ BoN-MAV          & $\mathbf{+3.33}$ [$+0.66$, $+6.00$] & $\mathbf{+2.33}$ [$+0.67$, $+4.67$] \\
Chess Val. $-$ cons      & $-0.33$ [$-2.00$, $+1.33$] & $\mathbf{+17.33}$ [$+12.67$, $+22.67$] \\
Chess Val. $-$ BoN-MAV   & $\mathbf{+1.33}$ [$+0.33$, $+3.00$] & $\mathbf{+8.00}$ [$+4.33$, $+12.67$] \\
Chess Opt. $-$ cons      & $\mathbf{+0.13}$ [$+0.03$, $+0.26$] & $\mathbf{+0.12}$ [$+0.04$, $+0.22$] \\
Chess Opt. $-$ BoN-MAV   & $\mathbf{+0.11}$ [$+0.05$, $+0.28$] & $\mathbf{+0.13}$ [$+0.04$, $+0.24$] \\
\bottomrule
\end{tabular}
\caption{\textbf{Paired differences with 95\% bootstrap CIs at $n = 8$.} Mean paired difference between APT@8 and each baseline, with a 95\% interval from a paired bootstrap over problems (10{,}000 resamples). Accuracy points except Chess Optimality (pawns). \textbf{Bold} marks intervals that strictly exclude zero.}
\label{tab:significance}
\end{table}

\subsection{Scaling the Number of Candidates}
\label{app:scaling}

\begin{figure}[h]
    \centering
    \includegraphics[width=0.8\linewidth]{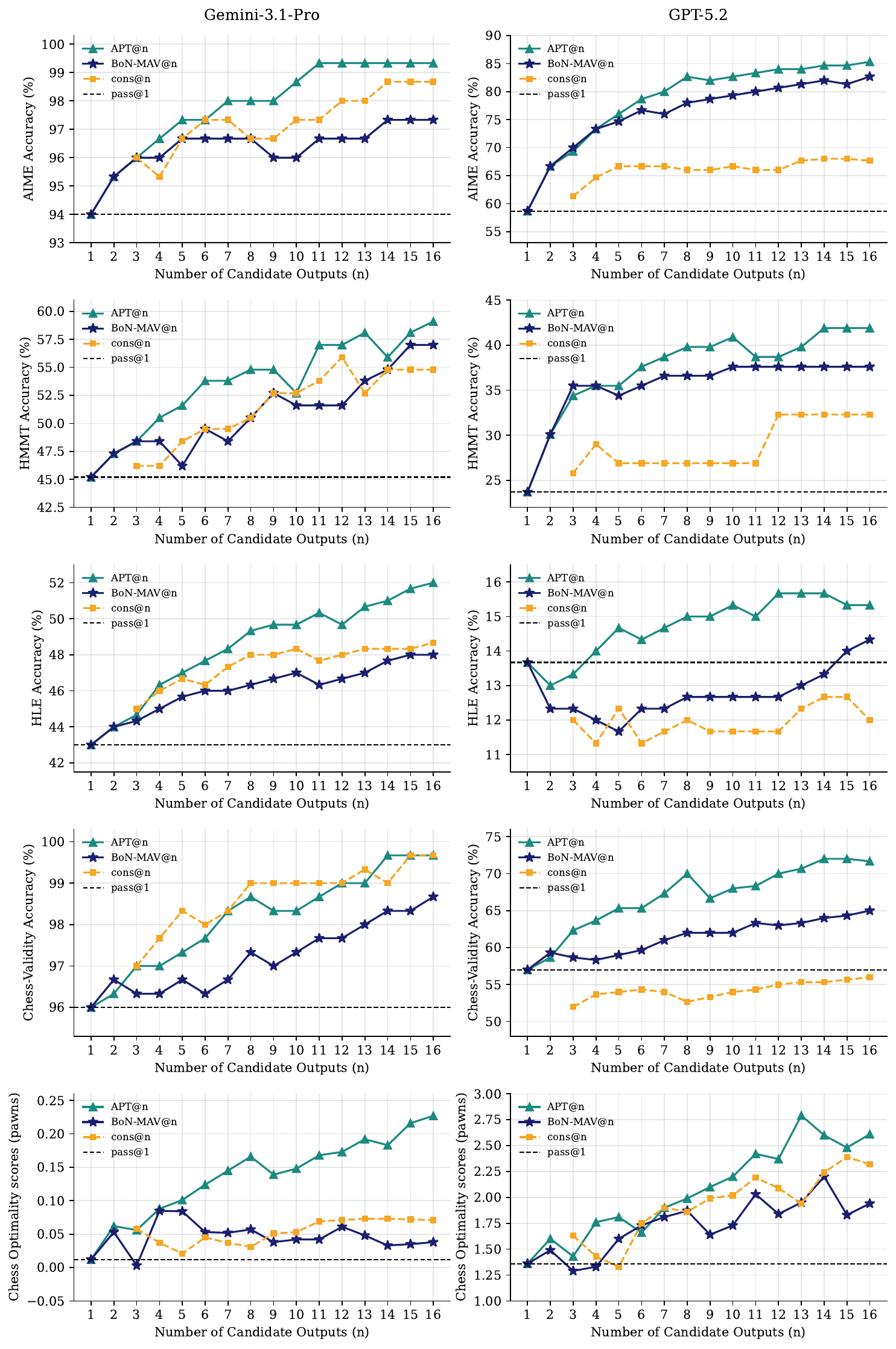}
    \caption{\textbf{Scaling curves for all methods.} Accuracy (pawn score for Chess Optimality) against the number of candidate outputs $n$, for Gemini-3.1-Pro (left) and GPT-5.2 (right) across all five datasets. Dashed line is pass@1. All three methods improve with $n$, but APT@$n$ is at or above both baselines over the full range in nearly every combination, and its margin widens with $n$ on Chess Optimality in particular.}
    \label{fig:scaling}
\end{figure}

Figure~\ref{fig:scaling} reports accuracy against the number of candidate outputs $n \in \{1, \dots, 16\}$ for APT, BoN-MAV and self-consistency, for both closed-source generators and all five datasets, with pass@1 shown as a reference line.  When $n$ is not a power of 2, the last candidate gets a first-round bye.

\subsection{Verifier-Count Ablation}
\label{app:verifier-ablation}

\begin{figure}[h]
    \centering
    \includegraphics[width=0.8\linewidth]{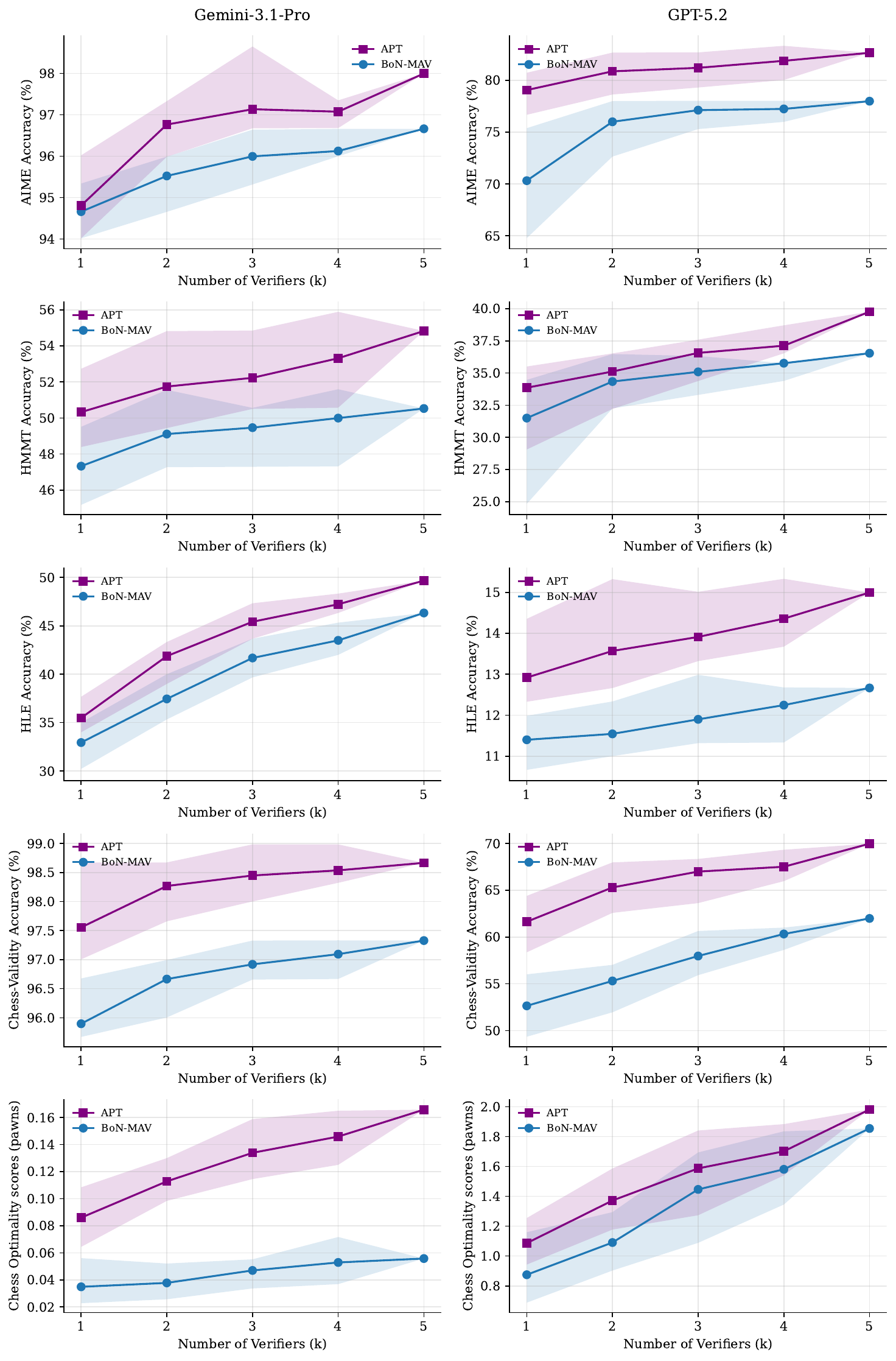}
    \caption{\textbf{Verifier-count ablation over all subsets.} Accuracy against the number of aspect verifiers $k$ for APT (purple) and BoN-MAV (blue), with shaded regions spanning all subsets of each size, for Gemini-3.1-Pro (left) and GPT-5.2 (right). Accuracy increases with $k$ for both methods on all five datasets, indicating that the aspects are complementary rather than redundant, and no single verifier accounts for the bulk of the performance. APT is at or above BoN-MAV at nearly every verifier count.}
    \label{fig:verifier-ablation}
\end{figure}

To test whether the aspect verifiers carry complementary signal, we remove verifiers from the full set one at a time and recompute accuracy for every resulting subset, for both APT and BoN-MAV. Figure~\ref{fig:verifier-ablation} plots accuracy against the number of verifiers $k$, with the shaded region spanning the subsets of each size.

\subsection{Bracket Sensitivity}
\label{app:bracket}

Single elimination commits to an outcome after only $n - 1$ matchups, so the bracket assignment could in principle matter a great deal. We sample five independent random seedings per problem and recompute the result in the same way; Table~\ref{tab:bracket} reports the mean across seedings with the observed range.

\begin{table}[h]
\centering
\small
\begin{tabular}{lcc}
\toprule
\textbf{Dataset} & \textbf{Gemini-3.1-Pro} & \textbf{GPT-5.2} \\
\midrule
AIME       & 98.13 [97.33, 99.33] & 82.68 [80.67, 84.00] \\
HMMT       & 55.26 [53.76, 58.06] & 38.06 [36.56, 40.86] \\
HLE        & 49.80 [49.00, 50.33] & 15.33 [14.33, 16.00] \\
Chess Val. & 98.27 [98.00, 99.00] & 70.60 [68.67, 71.33] \\
Chess Opt. & 0.171 [0.133, 0.206] & 2.048 [1.982, 2.104] \\
\bottomrule
\end{tabular}
\caption{\textbf{Sensitivity to bracket seeding.} Mean APT performance at $n = 8$ across five independent random bracket seedings per problem, with the range across seedings in brackets. Accuracy (\%) except Chess Optimality (pawns). Variation across seedings is small relative to the differences between methods in Table~\ref{tab:significance}, because most problems admit several correct candidates and a poor seeding for one of them does not remove the rest from the bracket.}
\label{tab:bracket}
\end{table}

\subsection{Quality of the Pairwise Signal}
\label{app:verifier-quality}

Three properties of the verifier preferences underpin the method: that a matchup between a correct and an incorrect solution is usually won by the correct one, that preferences are close to transitive, and that they are not dominated by presentation order.

\textbf{Per-matchup accuracy.} Table~\ref{tab:matchup-accuracy} reports how often the verifier majority prefers the correct solution when a correct and an incorrect candidate meet. Chess Optimality is omitted because it has no binary notion of correctness.

\begin{table}[h]
\centering
\small
\begin{tabular}{lcc}
\toprule
\textbf{Dataset} & \textbf{Gemini-3.1-Pro} & \textbf{GPT-5.2} \\
\midrule
AIME       & 89.5 & 85.6 \\
HMMT       & 85.2 & 82.9 \\
HLE        & 83.7 & 78.4 \\
Chess Val. & 92.3 & 87.9 \\
\bottomrule
\end{tabular}
\caption{\textbf{Per-matchup verifier accuracy (\%).} Rate at which the majority vote across aspect verifiers prefers the correct solution in a correct-versus-incorrect matchup, by dataset and generator (verifiers are Gemini-2.5-Flash and GPT-5-mini respectively).}
\label{tab:matchup-accuracy}
\end{table}

\textbf{Transitivity.} Tournament aggregation implicitly assumes preferences are not badly cyclic. We enumerate all triples of candidate solutions and measure the fraction for which the three pairwise preferences are transitive, separately for each aspect verifier. Table~\ref{tab:transitivity} shows that transitivity holds in 89-98\% of triples across all aspects and both verifier models. This rate is high, though not perfect, which is consistent with the Copeland ceiling being close to but slightly above single elimination (Appendix~\ref{app:tournament-formats}).

\begin{table}[h]
\centering
\small
\begin{tabular}{llcc}
\toprule
\textbf{Aspect} & \textbf{Strategy} & \textbf{Gemini-2.5-Flash} & \textbf{GPT-5-mini} \\
\midrule
General Correctness & Summarize         & 95.7 & 94.6 \\
General Correctness & Identify Mistakes & 97.3 & 95.7 \\
Domain Knowledge    & Direct            & 94.8 & 92.7 \\
General Correctness & Diff Comparison   & 95.7 & 93.5 \\
Edge Cases          & Direct            & 93.2 & 89.2 \\
Unit Conversions    & Step-by-Step      & 96.4 & 94.3 \\
Logical Soundness   & Step-by-Step      & 97.5 & 95.4 \\
Factual Accuracy    & Step-by-Step      & 96.2 & 95.7 \\
\bottomrule
\end{tabular}
\caption{\textbf{Transitivity of verifier preferences (\%).} Fraction of candidate triples for which the three pairwise preferences produced by a single aspect verifier are transitive.}
\label{tab:transitivity}
\end{table}

\textbf{Position bias.} Table~\ref{tab:position-bias} reports the raw rate at which the first-presented candidate (A) is chosen over the second (B), before the order-swapping correction described in Section~\ref{sec:pairwise-verification}. Most settings fall below 0.5, meaning the verifiers mildly favour the \emph{second} candidate rather than the first; the exceptions are Qwen3-8B solutions on HMMT and HLE. Either direction is a bias, which is why we query both orderings and discard order-dependent judgments rather than relying on the bias being small.

\begin{table}[h]
\centering
\small
\begin{tabular}{lccc}
\toprule
\textbf{Dataset} & \textbf{Gemini-3.1-Pro} & \textbf{GPT-5.2} & \textbf{Qwen3-8B} \\
\midrule
AIME       & 0.4340 & 0.3215 & 0.4739 \\
HMMT       & 0.4919 & 0.3482 & 0.6821 \\
HLE        & 0.4802 & 0.3348 & 0.5368 \\
Chess Val. & 0.4926 & 0.3544 & 0.4614 \\
Chess Opt. & 0.4380 & 0.3581 & 0.4564 \\
\bottomrule
\end{tabular}
\caption{\textbf{First-position selection rate.} Fraction of single-ordering comparisons in which the first-presented candidate is preferred, by dataset and by the generator whose solutions are being compared. A value of 0.5 indicates no positional preference.}
\label{tab:position-bias}
\end{table}

\subsection{Failure Analysis}
\label{app:failure}

Not every error APT makes is a selection error. We separate \emph{generator-limited} failures, where no correct candidate was sampled and no verification method could have succeeded, from \emph{selection} failures, where a correct candidate was present in the pool but was not chosen. Only the latter are attributable to APT. Table~\ref{tab:failure} reports the breakdown at $n = 16$. Chess Optimality is omitted as it has no binary correctness criterion.

\begin{table}[h]
\centering
\small
\begin{tabular}{llccc}
\toprule
\textbf{Dataset} & \textbf{Generator} & \textbf{APT correct} & \textbf{No correct candidate} & \textbf{Selection failure} \\
\midrule
\multirow{2}{*}{AIME}       & Gemini-3.1-Pro & 99.3 & 0.67  & 0.0 \\
                            & GPT-5.2        & 85.3 & 13.3 & 1.3 \\
\midrule
\multirow{2}{*}{HMMT}       & Gemini-3.1-Pro & 59.1 & 34.4 & 6.5 \\
                            & GPT-5.2        & 41.9 & 57.0 & 1.1 \\
\midrule
\multirow{2}{*}{HLE}        & Gemini-3.1-Pro & 52.0 & 34.3 & 13.7 \\
                            & GPT-5.2        & 15.3 & 72.0 & 12.67 \\
\midrule
\multirow{2}{*}{Chess Val.} & Gemini-3.1-Pro & 99.7 & 0.3  & 0.0 \\
                            & GPT-5.2        & 72.3 & 19.0 & 8.7 \\
\bottomrule
\end{tabular}
\caption{\textbf{Failure breakdown (\% of problems).} Problems are partitioned into those APT answers correctly, those where the generator produced no correct candidate, and those where a correct candidate existed but APT selected a different one. Only the last column is attributable to the verification method.}
\label{tab:failure}
\end{table}

Across datasets, the majority of errors are generator-limited: on HMMT with GPT-5.2, for instance, 57\% of problems had no correct candidate in the pool at all, against 1.1\% where APT had a correct candidate available and passed it over. Inspecting the selection failures qualitatively, two causes recur. The first is a weak margin in the preference vote, where the verifier majority is narrow and falls on the wrong side; these are the matchups where the aspect verifiers disagree with each other. The second is unfavourable bracket placement, where a correct candidate is drawn against another strong candidate in an early round and eliminated before reaching the final. The second cause is bounded by the bracket ablation in Appendix~\ref{app:bracket}, which shows the effect is small in aggregate.


\subsection{Varying Tournament Structure}
\label{app:tournament-formats}

In this section, we present preliminary experiments comparing different tournament formats on the chess validity and chess optimality domains. These experiments were conducted using GPT-5.2 (with GPT-5-mini verifiers) and Gemini-3.1-Pro (with Gemini-2.5-Flash verifiers), with $n = 8$ candidates and $m = 5$ aspect verifiers.

We compare three pairwise tournament formats against the non-pairwise baselines (Self-Consistency@8 and BoN-MAV@8):
\begin{itemize}
    \item \textbf{Single Elimination.} Candidates are paired and the majority-vote winner of each matchup advances. Requires $n - 1$ matchups.
    \item \textbf{Double Elimination.} Each candidate must lose twice before being eliminated. Losers drop to a losers' bracket and may still win the tournament. Requires between $2(n-1)$ and $2n - 1$ matchups.
    \item \textbf{Swiss.} Candidates are paired in each round based on their current win-loss record. After $\lceil \log_2 n \rceil$ rounds, the candidate with the best record is selected. Requires $\frac{n}{2} \cdot \lceil \log_2 n \rceil$ matchups.
\end{itemize}

Tables~\ref{tab:tournament-chess-validity} and~\ref{tab:tournament-chess-optimality} present the results on chess validity and chess optimality, respectively.

\begin{table}[h]
\centering
\caption{\textbf{Tournament format comparison on Chess Validity.} Accuracy (\%) with $n = 8$ candidates.}
\label{tab:tournament-chess-validity}
\small
\begin{tabular}{lcc}
\toprule
\textbf{Method} & \textbf{GPT-5.2} & \textbf{Gemini-3.1-Pro} \\
\midrule
Pass@1              & 57.33 & 96.00 \\
Self-Consistency@8  & 52.67 & 99.00 \\
BoN-MAV@8           & 62.00 & 97.33 \\
\midrule
Single Elimination  & 70.00 & 98.67 \\
Double Elimination  & \textbf{72.33} & \textbf{99.00} \\
Swiss               & 70.33 & 98.00 \\
\bottomrule
\end{tabular}
\end{table}

\begin{table}[h]
\centering
\caption{\textbf{Tournament format comparison on Chess Optimality.} pawn score (higher is better) with $n = 8$ candidates.}
\label{tab:tournament-chess-optimality}
\small
\begin{tabular}{lcc}
\toprule
\textbf{Method} & \textbf{GPT-5.2} & \textbf{Gemini-3.1-Pro} \\
\midrule
Pass@1              & 1.378 & 0.018 \\
Self-Consistency@8  & 1.866 & 0.031 \\
BoN-MAV@8           & 1.856 & 0.056 \\
\midrule
Single Elimination  & \textbf{1.982} & 0.166 \\
Double Elimination  & 1.905 & \textbf{0.194} \\
Swiss               & 1.940 & 0.173 \\
\bottomrule
\end{tabular}
\end{table}

On chess validity, all three tournament formats produce a large improvement over the baselines for GPT-5.2. We see performance jumping from 62\% (BoN-MAV) to 70-72.33\%. With Gemini-3.1-Pro, as mentioned before, gains are limited due to the saturation of the benchmark. Double elimination achieves the highest accuracy on both models, though the margin over single elimination is small (0.33-2.33\%). Swiss is slightly better with GPT-5.2 but slightly worse with Gemini-3.1-Pro. 

On chess optimality, the pattern is similar. We see that all tournament formats substantially outperform BoN-MAV, particularly for Gemini-3.1-Pro where the jump is from 0.056 to 0.166-0.194 pawns. Double elimination again achieves the highest score for Gemini-3.1-Pro (0.194), while for GPT-5.2 single elimination slightly leads (1.982 vs.\ 1.905 for double elimination). The differences between tournament formats are small relative to the gap between any tournament format and the non-pairwise baselines.

We also conducted a diagnostic experiment on the GPT-5.2 chess validity results, running all $\binom{8}{2} = 28$ pairwise comparisons for each of the 300 problems (with 5 verifiers per matchup). This full round-robin allows us to compute the Copeland winner (the candidate with the most pairwise matchup wins) which serves as a more comparison-intensive reference method. The Copeland method achieves 75.67\%, compared to 70\% for single elimination and a pass@8 oracle of 80\%. This confirms that single elimination captures most of the signal available in the full pairwise preference matrix. Furthermore, when the majority vote across verifiers compares a correct solution against an incorrect one, it selects the correct solution 87.9\% of the time, explaining the strong performance of all tournament formats: with pairwise signal this reliable, the correct solution is unlikely to be eliminated regardless of bracket structure.

These results support our choice of single elimination as the default tournament format in APT. While double elimination offers a small additional margin in some settings, single elimination achieves comparable performance with fewer matchups, making it the more compute-efficient choice.

\begin{table*}[t]
\centering
\small
\begin{tabular}{p{0.14\textwidth}p{0.82\textwidth}}
\toprule
\textbf{Domain} & \textbf{Prompt} \\
\midrule
AIME / HMMT &
\texttt{You are a helpful assistant skilled in math problem-solving. (\{IF AIME:\} Always express your final numerical answer as a three digit integer. If your final numerical answer is less than three digits, fill in leading zeros. For example, if your final numerical answer is '38', express your answer as '038'.) Always end your solution with the final numerical answer in latex, using '\textbackslash boxed\{<answer>\}'. If there is no solution, reply with an empty boxed '\textbackslash boxed\{\}'.} \newline \texttt{Please solve the following math problem step by step. Think carefully about your solution.} \newline \texttt{QUESTION: \{problem\}} \newline \texttt{Provide your detailed solution below:} \\
\midrule
HLE &
\texttt{Answer the following multiple choice question. Think step by step before answering, and then output the answer in the format of "The answer is (X)", where X is the LETTER of the correct answer.} \newline \texttt{QUESTION: \{problem\}} \newline \texttt{Think step by step, then answer with EXACTLY "The answer is (X)", where X is the LETTER of the correct answer. Do not include the answer text itself, only the letter. After stating your answer, give a detailed explanation of why this answer is correct.} \\
\midrule
Chess Validity &
\texttt{Given the following (in-progress) chess game, please complete the notation for the last shown move by filling in the destination square. Begin your solution by stating the answer. State your answer EXACTLY in the form "The answer is (X)", where X follows the regex [a-h][1-8], for example "The answer is (e5)." After stating your answer, give a brief explanation of why answer is valid.} \newline \texttt{GAME: \{problem\}} \\
\midrule
Chess Optimality &
\texttt{You are an expert chess player. Given the following chess position in FEN notation, suggest the MOST OPTIMAL next move for White. Think step by step about the position, threats, and candidate moves before deciding. Then state your suggested move EXACTLY in the form "The answer is (MOVE)", where MOVE is in Standard Algebraic Notation (SAN). Examples: "The answer is (Nf3)", "The answer is (e4)", "The answer is (O-O)", "The answer is (Bxc6)". After stating your move, give a brief explanation of why this move is optimal.} \newline \texttt{POSITION (FEN): \{problem\}} \\
\bottomrule
\end{tabular}
\caption{\textbf{Generator prompts.} The system prompt used to sample candidate solutions for each domain. \texttt{\{problem\}} is replaced with the problem text.}
\label{tab:gen-prompts}
\end{table*}

\begin{table*}[t]
\centering
\small
\begin{tabular}{p{0.14\textwidth}p{0.82\textwidth}}
\toprule
\textbf{Domain} & \textbf{System Prompt + Input Format} \\
\midrule
AIME / HMMT &
\texttt{You are a critical verifier tasked with comparing two mathematical solutions. You will be presented with a question and two proposed solutions. Your job is to determine which solution is better based on the specified criteria. Follow the instructions.} \newline\newline \texttt{QUESTION: \{question\}} \newline \texttt{SOLUTION A: \{solution\_a\}} \newline \texttt{SOLUTION B: \{solution\_b\}} \\
\midrule
HLE &
\texttt{You are a critical verifier tasked with comparing two multiple-choice answers. You will be presented with a question and two proposed solutions. Your job is to determine which solution is better based on the specified criteria. Follow the instructions.} \newline\newline \texttt{QUESTION: \{question\}} \newline \texttt{SOLUTION A: \{solution\_a\}} \newline \texttt{SOLUTION B: \{solution\_b\}} \\
\midrule
Chess Validity / Chess Optimality &
\texttt{You are a critical verifier tasked with comparing solutions related to moves in the game of chess. You will be presented with a question and two proposed solutions. Your job is to determine which solution is better based on the specified criteria. Follow the instructions.} \newline\newline \texttt{QUESTION: \{question\}} \newline \texttt{SOLUTION A: \{solution\_a\}} \newline \texttt{SOLUTION B: \{solution\_b\}} \\
\bottomrule
\end{tabular}
\caption{\textbf{Pairwise verifier system prompts by domain.} The system-level prompt prepended to all pairwise verifier queries. \texttt{\{question\}}, \texttt{\{solution\_a\}}, and \texttt{\{solution\_b\}} are replaced with the problem text and the two candidate solutions.}
\label{tab:vera-system-prompts}
\end{table*}

\begin{table*}[t]
\centering
\small
\begin{tabular}{p{0.19\textwidth}p{0.77\textwidth}}
\toprule
\textbf{Aspect--Strategy} & \textbf{Comparison Instruction} \\
\midrule
Unit Conversions / Step-by-Step &
\texttt{Compare how well each solution handles units in calculations. Which solution has better unit handling? Think out loud.} \\
\midrule
General Correctness / Summarize &
\texttt{Compare the overall quality and clarity of both solutions. Which solution is more comprehensive and well-explained? Think out loud.} \\
\midrule
General Correctness / Identify Mistakes &
\texttt{Compare both solutions for common mistakes and calculation errors. Which solution has fewer mistakes? Think out loud.} \\
\midrule
Domain Knowledge / Direct &
\texttt{Compare how well each solution applies domain knowledge and best practices. Which solution demonstrates better domain expertise? Think out loud.} \\
\midrule
Edge Cases / Direct &
\texttt{Compare how well each solution handles edge cases and boundary conditions. Which solution better addresses potential edge cases? Think out loud.} \\
\midrule
Logical Soundness / Step-by-Step &
\texttt{Compare the logical soundness of both solutions. Which solution follows more logical reasoning? Think out loud.} \\
\midrule
General Correctness / Diff Comparison &
\texttt{Compare the difficulty and sophistication of both solutions. Which solution demonstrates better problem-solving approach? Think out loud.} \\
\midrule
Factual Accuracy / Step-by-Step &
\texttt{Compare the factual accuracy of both solutions step by step. Which solution has more accurate facts? Think out loud.} \\
\bottomrule
\end{tabular}
\caption{\textbf{Pairwise verifier aspect prompts.} The aspect-specific instruction appended after the system prompt and input for each verifier. All prompts end with the format instruction: \texttt{To indicate your final answer, you MUST end your response with `FINAL PAIRWISE PREFERENCE: A' if Solution A is better, `FINAL PAIRWISE PREFERENCE: B' if Solution B is better, or `FINAL PAIRWISE PREFERENCE: NO PREFERENCE' if both solutions are equally good or equally flawed.} This suffix is omitted above for brevity.}
\label{tab:vera-aspect-prompts}
\end{table*}

\end{document}